\documentclass[sigconf]{acmart}
\AtBeginDocument{%
  }

\usepackage{tabularx}
\usepackage{multirow}
\usepackage{multicol}
\usepackage{url}
\usepackage{booktabs}
\usepackage{xcolor}
\usepackage{colortbl}
\usepackage{algorithm}
\usepackage{algorithmic}
\usepackage{tcolorbox}
\usepackage{pifont}
\usepackage{relsize}

\newcommand{\ours}{\texttt{APEER}\,}
\definecolor{lightgreen}{rgb}{0.88, 1, 0.88}
\copyrightyear{2025}
\acmYear{2025}
\setcopyright{cc}
\setcctype{by}
\acmConference[WWW Companion '25]{Companion Proceedings of the ACM Web Conference 2025}{April 28-May 2, 2025}{Sydney, NSW, Australia}
\acmBooktitle{Companion Proceedings of the ACM Web Conference 2025 (WWW Companion '25), April 28-May 2, 2025, Sydney, NSW, Australia}
\acmDOI{10.1145/3701716.3717574}
\acmISBN{979-8-4007-1331-6/2025/04}

\begin{document}

\title{\ours: \underline{A}utomatic \underline{P}rompt \underline{E}ngineering \underline{E}nhances Large Language Model \underline{R}eranking}
\author{Can Jin}
\authornote{Both authors contributed equally to this research.}
\email{can.jin@rutgers.edu}
\affiliation{%
  \institution{Rutgers University}
  \city{Piscataway}
  \state{New Jersey}
  \country{USA}
}

\author{Hongwu Peng}
\authornotemark[1]
\email{hongwu.peng@uconn.edu}
\affiliation{%
  \institution{University of Connecticut}
  \city{Storrs}
  \state{Connecticut}
  \country{USA}
}

\author{Shiyu Zhao}
\email{sz553@rutgers.edu}
\affiliation{%
  \institution{Rutgers University}
  \city{Piscataway}
  \state{New Jersey}
  \country{USA}
}

\author{Zhenting Wang}
\email{zhenting.wang@rutgers.edu}
\affiliation{%
  \institution{Rutgers University}
  \city{Piscataway}
  \state{New Jersey}
  \country{USA}
}

\author{Wujiang Xu}
\email{wujiang.xu@rutgers.edu}
\affiliation{%
  \institution{Rutgers University}
  \city{Piscataway}
  \state{New Jersey}
  \country{USA}
}

\author{Ligong Han}
\email{ligong.han@rutgers.edu}
\affiliation{%
  \institution{Rutgers University}
  \city{Piscataway}
  \state{New Jersey}
  \country{USA}
}

\author{Jiahui Zhao}
\email{jiahui.zhao@uconn.edu}
\affiliation{%
  \institution{University of Connecticut}
  \city{Storrs}
  \state{Connecticut}
  \country{USA}
}

\author{Kai Zhong}
\email{kaizhong@gmail.com}
\affiliation{%
  \institution{Independent}
  \city{Palo Alto}
  \state{California}
  \country{USA}
}

\author{Sanguthevar Rajasekaran}
\email{sanguthevar.rajasekaran@uconn.edu}
\affiliation{%
  \institution{University of Connecticut}
  \city{Storrs}
  \state{Connecticut}
  \country{USA}
}

\author{Dimitris N. Metaxas}
\email{dnm@cs.rutgers.edu}
\affiliation{%
  \institution{Rutgers University}
  \city{Piscataway}
  \state{New Jersey}
  \country{USA}
}
\renewcommand{\shortauthors}{Can Jin et al.}


\begin{abstract}
  Large Language Models (LLMs) have significantly enhanced Information Retrieval (IR) across various modules, such as reranking. Despite impressive performance, current zero-shot relevance ranking with LLMs heavily relies on human prompt engineering. Existing automatic prompt engineering algorithms primarily focus on language modeling and classification tasks, leaving the domain of IR, particularly reranking, underexplored. Directly applying current prompt engineering algorithms to relevance ranking is challenging due to the integration of query and long passage pairs in the input, where the ranking complexity surpasses classification tasks. To reduce human effort and unlock the potential of prompt optimization in reranking, we introduce a novel automatic prompt engineering algorithm named \ours. \ours iteratively generates refined prompts through feedback and preference optimization. Extensive experiments with four LLMs and ten datasets demonstrate the substantial performance improvement of \ours over existing state-of-the-art (SoTA) manual prompts. Furthermore, we find that the prompts generated by \ours exhibit better transferability across diverse tasks and LLMs.
\end{abstract}

\ccsdesc[500]{Information systems~Information retrieval}
\keywords{Prompt engineering, Information Retrieval, Large Language Model, ReRanking}


\maketitle

\section{Introduction}
Large Language Models (LLMs) have revolutionized the field of Natural Language Processing (NLP), achieving success across a variety of tasks \citep{achiam2023gpt,brown2020language,touvron2023llama,lyu2023attention}. One of the most impactful applications of LLMs is in Information Retrieval (IR), which focuses on efficiently retrieving information relevant to user queries \citep{hou2024large, fan2023recommender, xi2023towards}. Due to their advanced linguistic understanding and world knowledge, LLMs enhance IR systems in multiple modules, thereby attracting increasing interest \citep{liang2022holistic,qin2023large,sun2023chatgpt}.

Relevance ranking, which aims to rank a set of candidate passages by their relevance to a given query, is the most critical problem in IR \citep{fan2022pre}. Recently, a series of works have explored manual prompt approaches for LLM zero-shot reranking \citep{sun2023chatgpt, pradeep2023rankzephyr, ma2023zero}. The key challenge in prompting lies in the design of the prompt, which has emerged as a crucial technique known as prompt engineering \citep{brown2020language,nye2022show,yao2022react,prasad2023grips}. Despite the impressive results in reranking, manual prompt engineering typically requires substantial human effort and expertise, with subjective and limited guidelines.

\begin{figure}[!t]
    \centering
    \includegraphics[width=1\columnwidth]{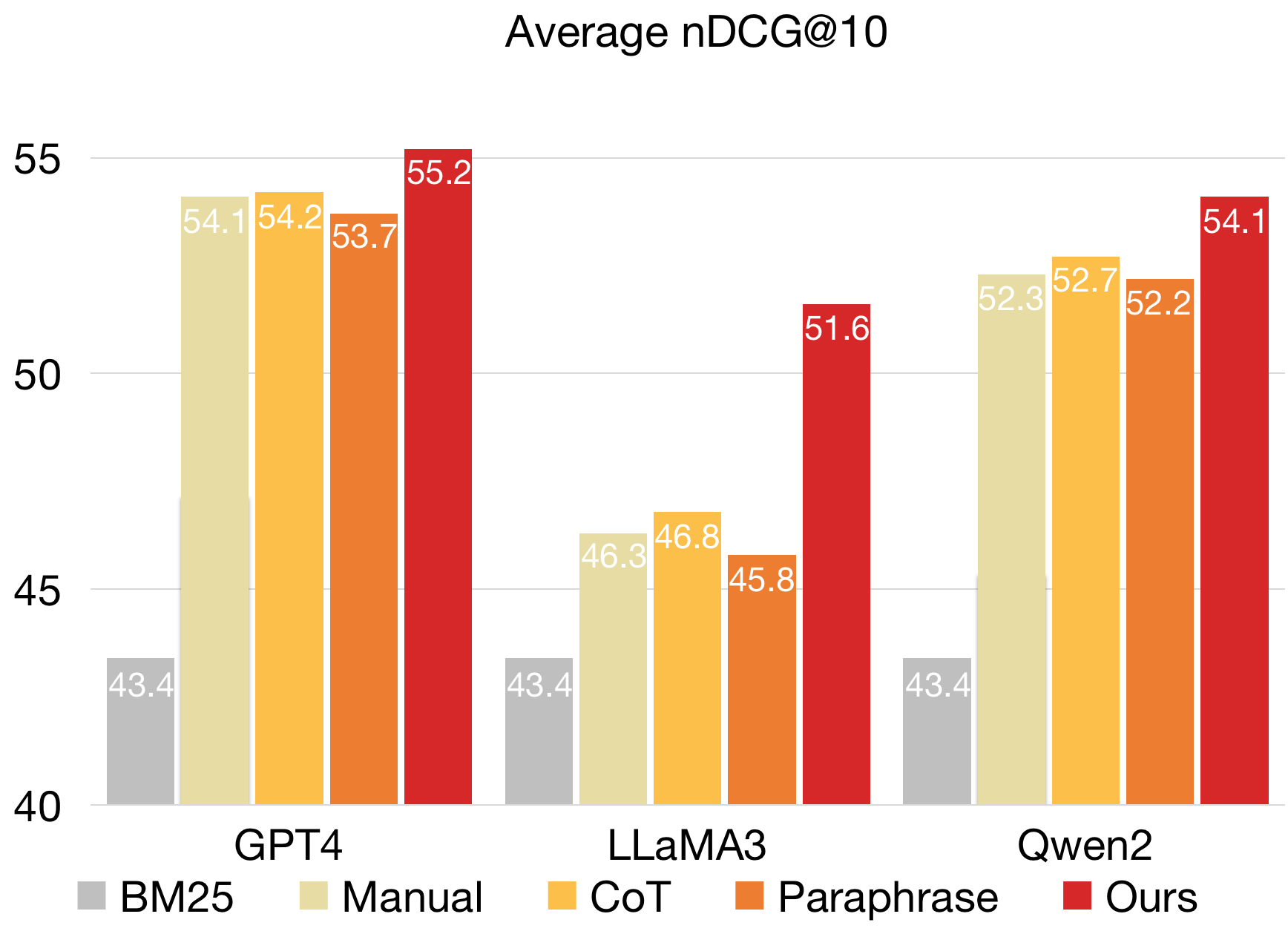}
    \caption{Performance overview of four prompting methods on GPT4, LLaMA3~\citep{llama3modelcard} and Qwen2~\citep{qwen2} models and BEIR datasets~\citep{thakur2021beir}. The manual prompt is RankGPT~\citep{sun2023chatgpt}. Modifying the manual prompt with CoT and paraphrasing yields marginal gains. 
    }
    \label{figure_illustration}
\end{figure}

Automatic prompt engineering can generate and select prompts autonomously, thereby reducing the human effort involved in prompt design and achieving impressive performance across various tasks such as language modeling and classification \citep{pryzant2023automatic, zhou2022large, guo2023connecting,liu2024enhanced}. However, the impact of automatic prompt engineering in the IR domain, particularly for zero-shot passage relevance ranking, has been less studied. Relevance ranking, which integrates a group of long passages into the input, presents unique challenges compared to language modeling and classification, and current automatic prompt engineering methods are sub-optimal in this field due to several reasons:
\ding{182} The input-output demonstrations for relevance ranking are more complex than those for language modeling. The input consists of query and passage pairs, and the output may not be unique, as various relevance ranks can serve as answers for a group of passages. \ding{183} The optimization process for relevance ranking is more challenging. It requires not only comprehension of the query but also comparison and relevance ranking of the passages. 
To this end, we aim to systematically address the following problem:
\begin{center}
\textit{How to design an automatic prompt optimization algorithm for passage relevance ranking?}
\end{center}

To answer the research question, we introduce \ours (\textbf{A}utomatic \textbf{P}rompt \textbf{E}ngineering \textbf{E}nhances LLM \textbf{R}eranking), which iteratively refines prompts through feedback generation and preference optimization. \ours comparison with state-of-the-art (SoTA) manual prompts in RankGPT~\citet{sun2023chatgpt}, chain-of-thought (CoT) prompting and paraphrasing is given in Figure~\ref{figure_illustration}. Results shows that \ours demonstrates significant improvement. In summary, 
our contributions are as follows:
\begin{itemize}
\item [$\star$] We investigate the effect of directly modifying current SoTA prompts using CoT and paraphrasing in relevance ranking and find their inefficacy in improving the performance of well-designed prompts.

\item [$\star$] To reduce human efforts and unlock the potential of prompt optimization, we propose a novel automatic prompt engineering algorithm, termed \ours, which generates refined prompts through feedback optimization and preference optimization to address the aforementioned challenges.

\item [$\star$] We conduct extensive experiments across diverse datasets and architectures, including newly released LLaMA3 and Qwen2. Empirical results consistently highlight the impressive performance advancements of \ours. For example, \ours achieves an average performance improvement of 5.29 (NDCG@10) on \textbf{eight} BEIR datasets compared to SoTA manual prompts on LLaMA3.

\item [$\star$] More interestingly, we demonstrate that the prompts generated by \ours exhibit enhanced transferability across multiple benckmarks and architectures.

\end{itemize}

\section{Related Works}
\subsection{Prompt Engineer}
Prompting offers a natural and intuitive interface for humans to interact with and utilize generalist models such as large language models (LLMs). Due to its flexibility, prompting has been widely adopted for various NLP tasks \citep{schick2021exploiting,brown2020language,sanh2021multitask,jin2024learning,wang2023brave,liuyue_GuardReasoner,liuyue_efficient_reasoning,wang2024large,wang2024ime,li2024human,shi2025explaining}. The chain-of-thought (CoT) prompting method was introduced to encourage LLMs to generate intermediate reasoning steps before arriving at a final answer \citep{kojima2022large, wei2022chain,wang2022self,jin2024impact}. However, LLMs require careful prompt engineering, whether manually \citep{reynolds2021prompt,jin2024exploring} or automatically \citep{pryzant2023automatic,zhou2022large,peng2023autorep}, due to the model's sensitivity \citep{jiang2020can,zhao2021calibrate,lu2022fantastically,lyu2022study,xu2024text} and their inability to understand prompts in the same way humans do \citep{webson2022prompt,lu2022fantastically,liu2024particle}. While many successful prompt tuning methods optimize over a continuous space using gradient-based techniques \citep{liu2023gpt,qin2021learning,lester2021power,jin2025visual,jin2025lorvp}, this becomes less practical at scale, as computing gradients becomes increasingly expensive and access to models shifts to APIs that may not provide gradient access. Another line of work focuses on discrete prompt search methods, such as prompt generation \citep{pryzant2023automatic,zhou2022large,guo2023connecting,ye2023prompt}, prompt scoring \citep{davison2019commonsense}, and prompt paraphrasing \citep{jiang2020can,yuan2021bartscore,liu2022distributed}, to optimize instructions by searching directly in the natural language hypothesis space. Prompt optimization for reranking has been less studied; \citet{cho2023discrete} explores discrete prompt optimization for query generation rather than relevance ranking for groups of passages. In this paper, we follow the line of work in prompt generation and propose a novel automatic prompt engineering algorithm for passage relevance ranking.

\subsection{LLMs for Information Retrieval}
IR is crucial for many knowledge-driven NLP applications \citep{zhu2023large,karpukhin2020dense,qu2021rocketqa,wu2017sequential,cao2024rough,zhang2024cut,yu2024robust,yu2025mtlue,SPIDER_ICML25,Lorasculpt_CVPR25}. LLMs have demonstrated remarkable efficacy in IR tasks \citep{zhu2023large, sun2023chatgpt, pradeep2023rankzephyr,yanhui2024dog,zhao2023cross,zhao2023sequential,10.1145/3689091.3690087,Shi2024}. IR typically consists of an initial, cost-effective retriever followed by a sophisticated reranker to refine the results~\citep{ma2023zero, craswell2020overview, nogueira2019multi,liu2024news}. Traditional supervised reranking methods \citep{nogueira2020document, zhuang2023rankt5, pradeep2023rankzephyr} often rely on fine-tuning transformer-based models with extensive training data, such as MS MARCO \citep{bajaj2016ms}. Recent research has explored zero-shot relevance ranking with LLMs. These methods can be broadly categorized into synthetic data generation and relevance ranking. For synthetic data generation, \citet{muennighoff2022sgpt} generate text embeddings using GPT for dense retrieval, while \citet{gao2023precise,wang2023query2doc} generate pseudo-documents for retrieval. In relevance ranking, RG \citep{liang2022holistic} generates relevance proxy tokens for ranking, while PRP \citep{qin2023large} compares the relevancies of two documents for a given query. RankGPT \citep{sun2023chatgpt} employs a zero-shot permutation generation method to reorder document relevance collectively and achieve improvements than RG and PRP using GPT4.

\begin{figure*}
    \centering
    \includegraphics[width=0.95\linewidth]{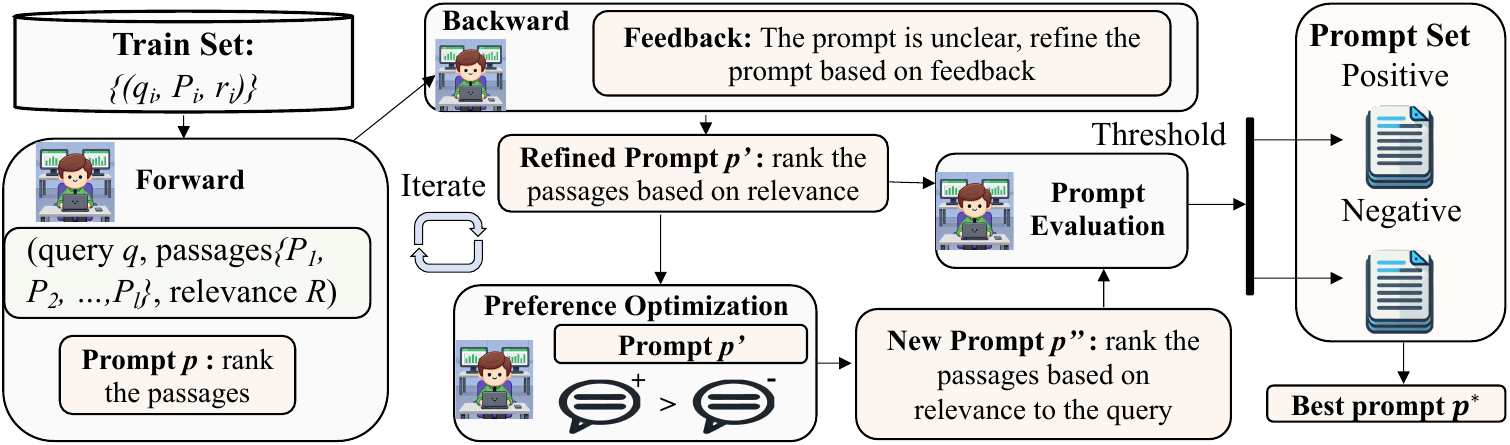}
    \caption{Overview of \ours. \ours iteratively refines prompts through two optimization steps. In Feedback Optimization, it refines the current prompt $p$ and creates a refined prompt $p'$ based on feedback. In Preference Optimization, it further optimizes $p'$ by learning preferences from a set of positive and negative prompt demonstrations.}
    \label{figure_method}
\end{figure*}

\section{Method}
In \ours, we attain superior prompts through two main optimization steps: \ding{182} Feedback optimization: we infer the current prompt, gather feedback on how to refine it, and then create a refined prompt based on the feedback. \ding{183} Preference optimization: we further optimize the refined prompt by learning preferences through a set of positive and negative prompt demonstrations. An overview of the training process of \ours is presented in Figure \ref{figure_method}.

\subsection{Problem Formulation}
IR is often implemented as a two-stage pipeline composed of a first-stage retriever and a second-stage reranker \citep{craswell2020overview}. For a given query $q$ sampled from a query distribution $\mathcal{Q}$, the retriever, such as BM25, efficiently returns a list of $l$ candidate passages $\mathcal{P} = \{P_1, P_2, \ldots, P_l\}$ from the original corpus $\mathcal{D}$ that are most relevant to $q$. The reranker then refines the relevance order of $\mathcal{P}$ to $q$ by further reranking the list of $l$ candidates according to either the same or a different metric used by the retriever. In \ours, we focus on improving the second-stage reranking performance with a fixed retriever, formulating the \textit{reranking optimization problem} as:
\begin{equation}
    \text{max} \ \mathbb{E}_{(q, \mathcal{P}, r) \in (\mathcal{Q}, \mathcal{D}, \mathcal{R})} \mathcal{M}(f([q, \mathcal{P}]; p), r),
    \label{formula_reranking_optimization_problem}
\end{equation}
where $\mathcal{R}$ is the standard relevance mapping set, $r \in \mathcal{R}$ indicates the standard relevance order between the query $q$ and passages $\mathcal{P}$, $f$ is an LLM, $\mathcal{M}$ is a predefined metric, and $p$ is the text prompt that will be concatenated with $q$ and $\mathcal{P}$ during inference. In our experiments, we choose normalized Discounted Cumulative Gain (nDCG) as the default metric for $\mathcal{M}$.

\subsection{Build Training Dataset}\label{section_build_training_dataset}
Corpus datasets in passage reranking are typically extremely large (see Table \ref{table_appendix_dataset_information} for corpus dataset information), with one corpus potentially containing hundreds of millions of tokens. Thus, directly utilizing all queries and corpus in current benchmarks as the training dataset would be enormously expensive. To build the training dataset $\mathcal{D}_{train}$, we first randomly sample a subset of queries from the standard training split in current benchmarks, such as the MS MARCO v1 training split \citep{bajaj2016ms}. For each sampled query $q$, using the standard relevance mapping set $\mathcal{R}$, we identify up to 10 positively relevant passages with a relevance score greater than zero and add them to the candidate passages set $\mathcal{P}$ for query $q$. To find negatively relevant passages, we use BM25 to retrieve the top 100 candidate passages most relevant to $q$. We then select the top passages with a relevance score of zero to $q$ and add them to $\mathcal{P}$. The final size of $\mathcal{P}$ for each query is 20. We then randomly shuffle the passage order in $\mathcal{P}$ and record the relevance mapping $r$ between $q$ and $\mathcal{P}$. Finally, $(q, \mathcal{P}, r)$ is added to the training dataset $\mathcal{D}_{train} = \{(q_i, \mathcal{P}_i, r_i)\}_{i=1}^n$. Following the same procedure, we can build the validation dataset $\mathcal{D}_{val} = \{(q_i, \mathcal{P}_i, r_i)\}_{i=1}^m$.

\subsection{Prompt Initialization} 
Due to the infinitely large search space, finding the optimal prompts from scratch can be extremely difficult. In \ours, we construct two initialized prompt sets to guide our optimization procedure: a positive prompts set $\mathcal{H}_{pos}$ and a negative prompts set $\mathcal{H}_{neg}$. The positive prompts serve as preferred examples, while negative prompts serve as dispreferred examples in prompt training. 

\textbf{Positive Prompt Initialization}. A good choice is utilizing the current SoTA manual prompt as the initial positive prompt $p_{pos}$. Various manual prompts have been proposed in zero-shot passage reranking, such as pointwise \citep{sachan2022improving,liang2022holistic}, pairwise \citep{qin2023large}, and listwise \citep{sun2023chatgpt,ma2023zero}. In our experiments, we choose the manual prompt from RankGPT \citep{sun2023chatgpt} as it has been proven to achieve superior performance compared to others \citep{sun2023chatgpt}. Other methods for initialization include leveraging the LLM $f$ to generate prompts and paraphrasing the manual prompt. 

\textbf{Negative Prompt Initialization}. We leverage a pretrained LLM to generate some prompt examples and choose the prompt that performs poorly on the validation dataset $\mathcal{D}_{val}$ as the initial negative prompt $p_{neg}$.

Both the positive prompt $p_{pos}$ and the negative prompt $p_{neg}$ are then evaluated on all queries in $\mathcal{D}_{val}$ to determine their performance. The positive prompt is then initialized as the current prompt $p = p_{init} = p_{pos}$. After initialization, we obtain the following:
\begin{equation}
    \begin{aligned}
        &\mathcal{H}_{pos} = \{p_{pos}\}, \\
        &\mathcal{H}_{neg} = \{p_{neg}\},
    \end{aligned}
    \label{formula_prompt_initialization}
\end{equation}

\subsection{Feedback Optimization} 

To update the current prompt $p$ and obtain refined prompts, we first infer it on a batch of data $\mathcal{B}=\{(q_i, \mathcal{P}_i, r_i)\}_{i=1}^k$ using the LLM $f$ and obtain the responses $S = \{s_i\}_{i=1}^k$, which constitutes the `forward' pass:
\begin{equation}
    s_i = f([q_i, \mathcal{P}_i]; p)
    \label{formula_forward}
\end{equation}

To attain the `gradient' ( i.e., feedback) on $\mathcal{B}$, we utilize the LLM $f$ to generate high-quality feedback on the current prompt based on the queries, passages, responses, and the relevance mapping:
\begin{equation}
    b_i = f([p, q_i, \mathcal{P}_i, s_i, r_i]; c_{fb}),
    \label{formula_feedback_generation}
\end{equation}
where $b_i$ is the feedback and $c_{fb}$ is the meta prompt for feedback generation.

To apply the obtained gradients to the current prompt, we `backward' $p$ by prompting the LLM to generate a refined prompt based on the feedback:
\begin{equation}
    p' = f([p, \{b_i\}_{i=1}^{k}]; c_{g}),
    \label{formula_backward}
\end{equation}
where $p'$ is the refined prompt and $c_{g}$ is the meta prompt for prompt refinement.

The refined prompt $p'$ is then evaluated on the validation dataset $\mathcal{D}_{val}$. If it achieves higher performance than $p_{init}$, it will be added to $\mathcal{H}_{pos}$; otherwise, it will be added to $\mathcal{H}_{neg}$.

\subsection{Preference Optimization}
Direct Preference Optimization \cite{rafailov2024direct} and Reinforcement Learning with Human Feedback (RLHF)~\cite{ouyang2022training} are prevalent techniques for steering the model's output towards the high-quality (potentially infrequent) responses within its training dataset. Within the framework of \ours, we have cataloged a collection of potential positive and negative responses within $\mathcal{H}_{pos}$ and $\mathcal{H}_{neg}$, respectively. Our objective is to refine the prompt $p'$ such that it is biased towards the optimal prompt contained in $\mathcal{H}_{pos}$. To achieve this, we employ a methodology where each refined prompt $p'$ is aligned with a high-quality prompt in $\mathcal{H}_{pos}$, utilizing pairs of positive and negative prompts $(p_{pos}, p_{neg})$ for demonstration purposes. We meticulously choose the top $t$ positive prompts from $\mathcal{H}_{pos}$ and the bottom $t$ prompts from $\mathcal{H}_{neg}$ to serve as our demonstration pairs. This procedure generates a new prompt $p''$ that exhibits a preference for positive prompts while avoiding negative ones:
\begin{equation}
    p'' = f([p', \{(p_{pos}, p_{neg})\}]; c_{pre}),
    \label{formula_preference_optimization}
\end{equation}
where $c_{pre}$ denotes the meta prompt for optimizing prompt preferences.

Subsequently, the performance of the newly generated prompt $p''$ relative to the baseline initialization prompt $p_{init}$ on the validation dataset $\mathcal{D}_{val}$ determines its categorization into either the positive prompt set $\mathcal{H}_{pos}$ or the negative prompt set $\mathcal{H}_{neg}$. Ultimately, the generation of prompts through both feedback optimization and preference optimization will maintain a balanced ratio of 1:1.

The algorithmic foundation of \ours is thoroughly outlined in Algorithm \ref{algorithm_ours}. Conceptually, the Feedback Optimization acts as a local optimizer for the current batch $\mathcal{B}$, whereas the Preference Optimization mechanism extends this local optimization by globally aligning the local optimized prompts towards superior global prompts, as identified in $\mathcal{H}_{pos}$, via preference learning from both positive and negative prompts across the dataset. The efficacy of Preference Optimization in enhancing the quality of prompts is evidenced by our ablation study, presented in Table \ref{table_ablation_preference_optimization}.

\begin{algorithm}[ht]
\caption{\ours}
\label{algorithm_ours}
\begin{algorithmic}
\STATE {\bfseries Input:} LLM $f$, Training Dataset $\mathcal{D}_{train} = \{(q_i, \mathcal{P}_i, r_i)\}_{i=1}^n$, Validation Dataset $\mathcal{D}_{val} = \{(q_i, \mathcal{P}_i, r_i)\}_{i=1}^m$, $p = p_{init} = p_{pos}$, Positive Prompts History $\mathcal{H}_{pos} = \{p_{pos}\}$, Negative Prompts History $\mathcal{H}_{neg} = \{p_{neg}\}$, Meta Prompts $c_{fb}$, ${c_g}$, $c_{pre}$.
\FOR{$e = 1$ to Epochs}
\STATE Sample a batch of data $\mathcal{B}\subset \mathcal{D}_{train}$
\STATE \textbf{Initialization}: $p$ is best prompt in $\mathcal{H}_{pos}$ 
\STATE \textbf{Feedback Optimization}: \\
\quad Forward: Response $s_i = f([q_i, \mathcal{P}_i]; p)$ \\
\quad Backward:\\
\quad \quad   Feedback $b_i = f([p, q_i, \mathcal{P}_i, s_i, r_i]; c_{fb})$\\
\quad \quad   $p' = f([p, \{b_i\}_{i=1}^{k}]; c_{g})$
\STATE Evaluate and add $p'$ to $\mathcal{H}_{pos}$ or $\mathcal{H}_{neg}$
\STATE \textbf{Preference Optimization}:\\
\quad $p'' = f([p', \{(p_{pos}, p_{neg})\}]; c_{pre})$
\STATE Evaluate and add $p''$ to $\mathcal{H}_{pos}$ or $\mathcal{H}_{neg}$
\ENDFOR
\RETURN The best prompt $p^* \in \mathcal{H}_{pos}$
\end{algorithmic}
\end{algorithm}

\section{Experiments}
To evaluate the effectiveness of \ours, we adhere to the standard reranking evaluation methodology. Specifically, we assess the reranking performance of the top 100 passages retrieved by a first-stage retriever, such as BM25, using Pyserini\footnote{\url{https://github.com/castorini/pyserini}}. Additionally, we conduct extensive experiments to: (1) demonstrate the superior performance of \ours on in-domain tasks; (2) illustrate the transferability of \ours prompts to out-of-domain tasks; and (3) exhibit the transferability of \ours prompts across various architectures. Furthermore, we perform in-depth ablation studies to evaluate the impact of our novel preference optimization, as well as the effects of different training dataset sizes in \ours.

\subsection{Implementation Details}\label{implementation_details}
\paragraph{Models.} Our experiments utilize two closed-source models, GPT3.5-Turbo-0301 and GPT4-0613~\citep{achiam2023gpt}, as well as two open-source models, LLaMA3-70B~\citep{llama3modelcard} and Qwen2-72B~\citep{qwen2}.

\begin{table}[t]
\centering
\caption{Performance overview (nDCG@\{1,5,10\}) of \ours and baseline methods trained on GPT-4, LLaMA-3, and Qwen-2 with MS MARCO samples, evaluated on TREC-DL19 and TREC-DL20. \ours consistently outperforms the baselines. Manual refers to the RankGPT~\cite{sun2023chatgpt} baseline. The best performance for each model is marked in \textbf{bold}, while the overall best performance is highlighted in \colorbox{lightgreen}{green}.}
\label{table_superior_trec}
\renewcommand{\arraystretch}{1.1}
\resizebox{0.5\textwidth}{!}{
\begin{tabular}{l|l|ccc|ccc}
\toprule
\multirow{3}{*}[-1.5ex]{\textbf{Model}} & \textbf{Dataset} & \multicolumn{3}{c}{\textbf{TREC-DL19}} & \multicolumn{3}{c}{\textbf{TREC-DL20}} \\
\cmidrule(lr){2-8}
& \textbf{nDCG} & @1 & @5 & @10 & @1 & @5 & @10 \\
\cmidrule(lr){2-8}
& \textit{BM25} & 54.26 & 52.78 & 50.58 & 57.72 & 50.67 & 47.96 \\ 
\midrule
\multirow{4}{*}[-1.5ex]{\textbf{GPT4}} & \textit{Manual} & 80.62 & 77.83 & 74.89 & 79.73 & 73.15 & 70.14 \\ 
& \textit{CoT} & 81.01 & 78.04 & 75.20 & 80.25 & 74.13 & 70.42 \\ 
& \textit{Paraphrase}  & 81.01 & 78.47 & 74.76 & 80.13 & 74.23 & 70.01 \\ 
& \ours & \cellcolor{lightgreen}\textbf{84.11} & \cellcolor{lightgreen}\textbf{79.73} & \cellcolor{lightgreen}\textbf{76.22} & \cellcolor{lightgreen}\textbf{82.72} & \cellcolor{lightgreen}\textbf{75.88} & \cellcolor{lightgreen}\textbf{70.78} \\
\midrule
\multirow{4}{*}[-1.5ex]{\textbf{LLaMA3}} & \textit{Manual} & 76.35 & 74.86 & 71.89 & 79.11 & 70.53 & 67.37 \\ 
& \textit{CoT} & 77.52 & 74.96 & 71.83 & 79.94 & 71.83 & 68.32 \\ 
& \textit{Paraphrase}  & 74.81 & 74.60 & 71.79 & 78.09 & 69.73 & 66.84 \\ 
& \ours & \textbf{81.40} & \textbf{76.57} & \textbf{73.01} & \textbf{81.79} & \textbf{72.25} & \textbf{68.99} \\
\midrule
\multirow{4}{*}[-1.5ex]{\textbf{Qwen2}} & \textit{Manual} & 80.44 & 76.63 & 72.78 & 79.53 & 72.68 & 68.80 \\ 
& \textit{CoT} & 81.01 & 78.29 & 73.92 & 79.94 & 73.03 & 69.18 \\ 
& \textit{Paraphrase}  & 80.62 & 76.86 & 73.08 & 79.63 & 72.80 & 68.95 \\ 
& \ours & \textbf{83.33} & \textbf{79.21} & \textbf{75.11} & \textbf{81.17} & \textbf{73.43} & \textbf{69.78} \\
\bottomrule
\end{tabular}}
\end{table}

\begin{table*}[ht]
\centering
\caption{Performance overview (nDCG@10) of \ours trained on GPT4, LLaMA3, and Qwen2 using MS MARCO samples, and evaluated on eight BEIR datasets. \ours prompts consistently demonstrate superior performance compared to baselines when transferred to BEIR datasets. Manual refers to the RankGPT~\cite{sun2023chatgpt} baseline.}
\label{table_superior_transfer_beir}
\resizebox{1\textwidth}{!}{
\begin{tabular}{l|l|cccccccc|c}
\toprule
\multirow{2}{*}[-0.8ex]{\textbf{Model}} & \textbf{Dataset} & \textbf{Covid} & \textbf{NFCorpus} & \textbf{Signal} & \textbf{News} & \textbf{Robust04} & \textbf{Touche} & \textbf{DBPedia} & \textbf{SciFact} & \textbf{BEIR (Average)} \\
\cmidrule(lr){2-11}
& \textit{BM25} & 59.47 & 33.75 & 33.05 & 39.52 & 40.70 & \cellcolor{lightgreen}44.22 & 31.80 & 67.89 & 43.80 \\ 
\midrule
\multirow{4}{*}{\textbf{GPT4}} & \textit{Manual} & 83.98 & 38.83 & 33.90 & 52.82 & 59.74 & 40.72 & 47.12 & 75.61 & 54.09 \\ 
& \textit{CoT} & 85.51 & 38.33 & 33.45 & 51.90 & 59.92 & 40.45 & 47.53 & 76.08 & 54.15 \\ 
& \textit{Paraphrase}  & 84.15 & 38.76 & 33.60 & 50.46 & 59.35 & 40.72 & 47.19 & 75.26 & 53.69 \\ 
& \ours & \cellcolor{lightgreen}\textbf{86.09} & \cellcolor{lightgreen}\textbf{40.19} & \cellcolor{lightgreen}\textbf{34.08} & \cellcolor{lightgreen}\textbf{54.77} & \cellcolor{lightgreen}\textbf{60.15} & \textbf{40.91} &
\cellcolor{lightgreen}\textbf{48.06} & 
\cellcolor{lightgreen}\textbf{77.02} &
\cellcolor{lightgreen}\textbf{55.16} \\
\midrule
\multirow{4}{*}{\textbf{LLaMA3}} & \textit{Manual} & 76.15 & 34.95 & 33.29 & 42.11 & 47.38 & 30.54 & 45.40 & 60.72 & 46.32 \\ 
& \textit{CoT} & 77.46 & 35.49 & 33.37 & 42.37 & 47.96 & 30.83 & 45.59 & 60.91 & 46.75 \\ 
& \textit{Paraphrase}  & 74.54 & 34.59 & 33.12 & 41.63 & 47.04 & 29.23 & 45.26 & 60.56 & 45.75 \\ 
& \ours & \textbf{83.86} & \textbf{38.93} & \textbf{33.41} & \textbf{52.11} & \textbf{56.03} & \textbf{35.25} & \textbf{46.13} & \textbf{67.16} & \textbf{51.61} \\
\midrule
\multirow{4}{*}{\textbf{Qwen2}} & \textit{Manual} & 80.07 & 38.07 & 32.87 & 47.35 & 59.24 & 41.02 & 45.53 & 74.08 & 52.28 \\ 
& \textit{CoT} & 81.45 & 38.19 & 32.96 & 47.61 & 59.42 & 41.22 & 45.80 & 74.55 & 52.65 \\ 
& \textit{Paraphrase} & 79.72 & 38.03 & 32.86 & 47.13 & 59.13 & 40.92 & 45.50 & 73.89 & 52.15 \\ 
& \ours & \textbf{85.07} & \textbf{39.30} & \textbf{33.06} & \textbf{50.83} & \textbf{59.61} & \textbf{41.61} & \textbf{46.88} & \textbf{76.56} & \textbf{54.12} \\
\bottomrule
\end{tabular}}
\end{table*}

\paragraph{Benchmarks.} We evaluate the effectiveness of \ours on three benchmarks: TREC~\citep{craswell2020overview} and BEIR~\citep{thakur2021beir}, which collectively include ten datasets. \textbf{TREC} is a widely adopted benchmark in IR research. We use the test sets from the TREC-DL19 and TREC-DL20 competitions, both of which employed the MS MARCO v1 passage corpus. \textbf{BEIR} encompasses diverse retrieval tasks and domains. We select the test sets of eight tasks from BEIR to evaluate our approach: (i) Covid, which retrieves scientific articles for COVID-19-related questions; (ii) NFCorpus, a biomedical information retrieval dataset; (iii) Signal, which retrieves relevant tweets for a given news title; (iv) News, which retrieves relevant news articles for news headlines; (v) Robust04, which evaluates poorly performing topics; (vi) Touche, an argument retrieval dataset; (vii) DBPedia, which retrieves entities from the DBpedia corpus; and (viii) SciFact, which retrieves evidence for scientific claim verification.

\paragraph{Baselines.} We compare \ours to four baselines: (1) \textit{BM25}~\citep{lin2021pyserini}, which serves as a fundamental sanity check by directly using the ranking results from the first-stage retrieval; (2) \textit{Manual Prompt}, where we select the current state-of-the-art (SoTA) manual prompt, RankGPT~\citep{sun2023chatgpt}; (3) \textit{CoT}, which uses the manual prompt concatenated with "Let's think step by step" as the CoT prompt; and (4) \textit{Paraphrase}, where we utilize the LLM to paraphrase the manual prompt to obtain a paraphrased version. We choose listwise reranking as our default reranking method, as it achieves superior performance compared to pointwise and pairwise reranking~\citep{sun2023chatgpt,qin2023large}. The implementation details of baselines is shown in Appendix~\ref{appendix_implementation_details}.

\paragraph{Training and Evaluation.} 
We construct the training and validation datasets as described in Section \ref{section_build_training_dataset}, using queries sampled from the standard MS MARCO v1 training split \citep{bajaj2016ms}. The same dataset is utilized for both training and validation, with the default number of queries set at 100. The initialization of the positive prompt is based on the SoTA manual prompt from RankGPT \citet{sun2023chatgpt}. The negative prompt initialization is generated by the training models. Optimal hyperparameters are determined through grid search. We evaluate zero-shot performance using normalized Discounted Cumulative Gain (nDCG) at rank cutoffs of \{1,5,10\} (nDCG@\{1,5,10\}) and the results are averaged over three runs. It is important to note that we use the Azure API for the GPT4-0613 model, which differs from the GPT4-0314 model used in RankGPT~\citep{sun2023chatgpt}. Additionally, RankGPT utilizes GPT4 to rerank the top 30 passages initially reranked by GPT3.5 on BEIR. These differences result in discrepancies between the RankGPT results in our study and those reported in~\citep{sun2023chatgpt}. Further implementation details are available in Appendix~\ref{appendix_implementation_details}.

\begin{table*}[ht]
\centering
\caption{Performance overview (nDCG@10) of applying GPT4 and Qwen2 generated prompts on GPT3.5 and LLaMA3 models, and evaluated on two TREC-DL datasets and eight BEIR datasets. \ours prompts consistently demonstrate superior transferability across models, outperforming baseline methods. Manual refers to the RankGPT~\cite{sun2023chatgpt} baseline.}
\label{table_superior_model_transfer_beir}
\resizebox{1\textwidth}{!}{
\begin{tabular}{l|cc|cccccccc|c}
\toprule
\textbf{Dataset} & \textbf{DL19} & \textbf{DL20} & \textbf{Covid} & \textbf{NFCorpus} & \textbf{Signal} & \textbf{News} & \textbf{Robust04} & \textbf{Touche} & \textbf{DBPedia} & \textbf{SciFact} & \textbf{BEIR (Average)} \\
\midrule
\textit{BM25} & 50.58 & 47.96 & 59.47 & 33.75 & 33.05 & 39.52 & 40.70 & \cellcolor{lightgreen}44.22 & 31.80 & 67.89 & 43.80 \\ 
\midrule
\multicolumn{12}{c}{\textbf{GPT4 $\rightarrow$ GPT3.5}} \\ 
\midrule
\textit{Manual} & 65.80 & 62.91 & 76.67 & 35.62 & 32.12 & 48.85 & 50.62 & 36.18 & 44.47 & 70.43 & 49.37 \\ 
\textit{CoT} & 65.15 & 62.24 & 76.02 & 35.81 & 32.78 & 49.98 & 50.64 & 37.27 & 43.82 & 70.90 & 49.65 \\ 
\textit{Paraphrase}  & 64.86 & 61.74 & 74.28 & 35.16 & 31.08 & 48.60 & 50.34 & 37.11 & 43.42 & 70.11 & 48.76 \\ 
\ours & \textbf{67.47} & \textbf{63.29} & \cellcolor{lightgreen}\textbf{81.57} & \textbf{37.56} & \textbf{32.98} & \cellcolor{lightgreen}\textbf{50.44} & \textbf{52.77} & \textbf{39.48} & \textbf{44.67} & \cellcolor{lightgreen}\textbf{72.87} & \cellcolor{lightgreen}\textbf{51.54} \\
\midrule
\multicolumn{12}{c}{\textbf{Qwen2 $\rightarrow$ LLaMA3}} \\ 
\midrule
\textit{Manual} & 71.99 & 67.37 & 76.15 & 34.95 & 33.29 & 42.11 & 47.38 & 30.54 & 45.40 & 60.72 & 46.32 \\ 
\textit{CoT} & 71.83 & 68.32 & 77.46 & 35.49 & 33.37 & 42.37 & 47.96 & 30.83 & 45.59 & 60.91 & 46.75 \\ 
\textit{Paraphrase}  & 71.87 & 67.43 & 75.11 & 35.08 & 33.01 & 41.89 & 47.36 & 29.79 & 45.59 & 60.75 & 46.07 \\ 
\ours & \cellcolor{lightgreen}\textbf{72.65} & \cellcolor{lightgreen}\textbf{68.79} & \textbf{80.28} & \cellcolor{lightgreen}\textbf{38.85} & \cellcolor{lightgreen}\textbf{33.62} & \textbf{46.66} & \cellcolor{lightgreen}\textbf{55.71} & \textbf{36.02} & \cellcolor{lightgreen}\textbf{46.08} & \textbf{68.06} & \textbf{50.66} \\
\bottomrule
\end{tabular}}
\end{table*}

\subsection{Superior Performance}
\paragraph{In-domain Results.} To assess the effectiveness of \ours prompts on in-domain tasks, we apply \ours to GPT4, LLaMA3, and Qwen2 training on MS MARCO samples. The evaluation is conducted on TREC-DL19 and TREC-DL20, which also use the MS MARCO corpus. Several positive observations can be drawn from the results shown in Table \ref{table_superior_trec}: \ding{182} \ours is capable of generating superior prompts compared to baselines across diverse architectures, effectively enhancing the initialized manual prompts. For example, it achieves \{5.05, 1.71, 1.12\} higher nDCG@\{1, 5, 10\} on LLaMA3 and DL19 than manual prompts. While CoT can enhance the performance of manual prompts, \ours consistently outperforms CoT across all models and datasets. Moreover, direct paraphrasing of the manual prompts leads to inferior performance, underscoring the importance of prompt training. \ding{183} With \ours, a weaker model can sometimes achieve better performance than a stronger model. For instance, Qwen2 with \ours achieves \{2.71, 1.38, 0.22\} higher nDCG@\{1, 5, 10\} than GPT4 with manual prompts, further demonstrating the effectiveness of \ours. \ding{184} GPT4 with \ours achieves the best performance across all prompting methods, models, and datasets.

\paragraph{Transferability Across Datasets.} Superior prompts should be generalizable across different datasets. To investigate the transferability of \ours on out-of-domain tasks, we conduct experiments using \ours trained on MS MARCO samples and evaluate them on eight BEIR datasets, which feature more diverse types of queries and corpora compared to TREC-DL and MS MARCO. The results, presented in Table \ref{table_superior_transfer_beir}, reveal the following: \ding{182} \ours consistently achieves the best performance across eight BEIR datasets and three model architectures. Notably, \ours shows average nDCG@10 improvements of \{1.07, 5.29, 1.84\} over manual prompts for GPT4, LLaMA3, and Qwen2, respectively. This demonstrates the effectiveness of \ours prompts on out-of-domain datasets. \ding{183} Simple application of CoT and paraphrased prompts does not significantly improve over manual prompts, highlighting the superiority of \ours prompt training. \ding{184} With \ours prompts, Qwen2 even achieves higher performance than GPT4, further underscoring the significance of our method. The transferability of \ours across diverse datasets enhances its practicality in real-world applications.

\paragraph{Transferability Across Models.} We further investigate whether prompts trained on one model architecture using \ours can be transferred to models with different architectures. We apply the prompts obtained by \ours and baseline methods on GPT4 and Qwen2 to GPT3.5 and LLaMA3 models, respectively. The results on two TREC-DL datasets and four BEIR datasets are shown in Table \ref{table_superior_model_transfer_beir}. Several positive observations can be drawn: \ding{182} Prompts trained on a strong model can be transferred to a significantly weaker model. For example, when applying \ours prompts from GPT4 to GPT3.5, they consistently achieve better performance than all baselines on all TREC-DL and BEIR datasets. \ding{183} \ours prompts can transfer across models with comparable performance. For instance, prompts trained on Qwen2 achieve significant performance improvements over manual prompts when applied to LLaMA3. The transferability of \ours across different architectures further enhances its practical utility in real-world applications.

\subsection{In-depth Dissection of \ours}
\paragraph{Preference Optimization.} In \ours, we propose a novel Preference Optimization method based on preference learning from (positive prompt, negative prompt) demonstrations. To investigate the impact of Preference Optimization in \ours, we conduct experiments using LLaMA3 trained on MS MARCO samples, with and without Preference Optimization, while keeping all other configurations the same. We evaluate performance on two TREC datasets. The results, shown in Table \ref{table_ablation_preference_optimization}, reveal that: \ding{182} Preference Optimization is effective in \ours, as \ours with Preference Optimization achieves higher performance than \ours without it. \ding{183} \ours without Preference Optimization still produces better prompts than baselines, further indicating the overall effectiveness of \ours.

\begin{table}[t]
\centering
\caption{Ablation results of Preference Optimization in \ours. We train \ours with and without Preference Optimization (denoted as \ours w. PO and \ours w.o. PO, respectively) on MS MARCO samples using LLaMA3, and evaluate on TREC-DL19 and TREC-DL20.}
\label{table_ablation_preference_optimization}
\resizebox{0.5\textwidth}{!}{
\begin{tabular}{l|ccc|ccc}
\toprule
\textbf{Dataset} & \multicolumn{3}{c}{\textbf{TREC-DL19}} & \multicolumn{3}{c}{\textbf{TREC-DL20}} \\
\midrule
\textbf{nDCG} & @1 & @5 & @10 & @1 & @5 & @10 \\
\midrule
\textit{BM25} & 54.26 & 52.78 & 50.58 & 57.72 & 50.67 & 47.96 \\ 
\midrule
\textit{Manual} & 76.35 & 74.86 & 71.99 & 79.11 & 70.53 & 67.37 \\ 
\textit{CoT} & 77.52 & 74.96 & 71.83 & 79.94 & 71.83 & 68.32 \\ 
\textit{Paraphrase}  & 74.81 & 74.60 & 71.79 & 78.09 & 69.73 & 66.84 \\ 
\midrule
\ours w.o. PO  & 78.68 & 75.33 & 72.41 & 81.17 & 71.97 & 68.39 \\ 
\ours w. PO & \textbf{81.40} & \textbf{76.57} & \textbf{73.01} & \textbf{81.79} & \textbf{72.25} & \textbf{68.99} \\
\bottomrule
\end{tabular}}
\end{table}

\begin{figure}
    \centering
    \includegraphics[width=1\linewidth]{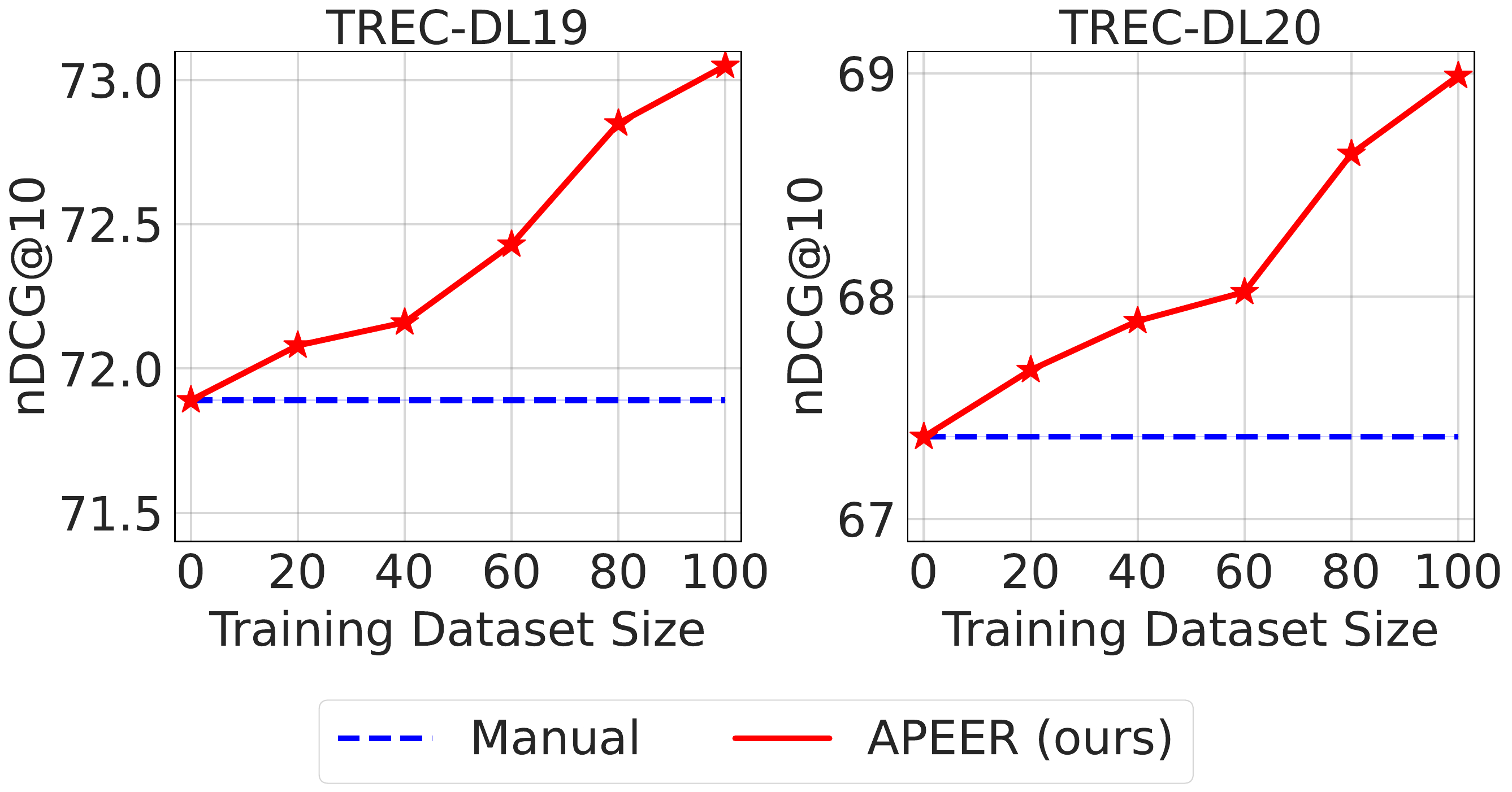}
        \caption{Ablation results of training dataset size. We train LLaMA3 model on various training dataset sizes and evaluate on TREC-DL19 and TREC-DL20.}
    \label{figure_ablation_training_dataset_size}
\end{figure}

\paragraph{Impact of Training Dataset Size.} We conduct experiments to investigate the influence of training dataset size on the performance of \ours. Following the procedure outlined in Section \ref{section_build_training_dataset}, we construct training datasets with varying numbers of queries. We then train the LLaMA3 model on these datasets using \ours, with the validation datasets being identical copies of the training datasets. The results on the TREC datasets, shown in Figure \ref{figure_ablation_training_dataset_size}, indicate that as the training dataset size increases, \ours achieves better performance. In our default setting, we utilize a training dataset size of 100 to attain superior prompts while maintaining moderate training costs. Further increasing the dataset size may improve performance, but it will also escalate training costs.

\begin{figure}[t]
    \centering
    \includegraphics[width=1\linewidth]{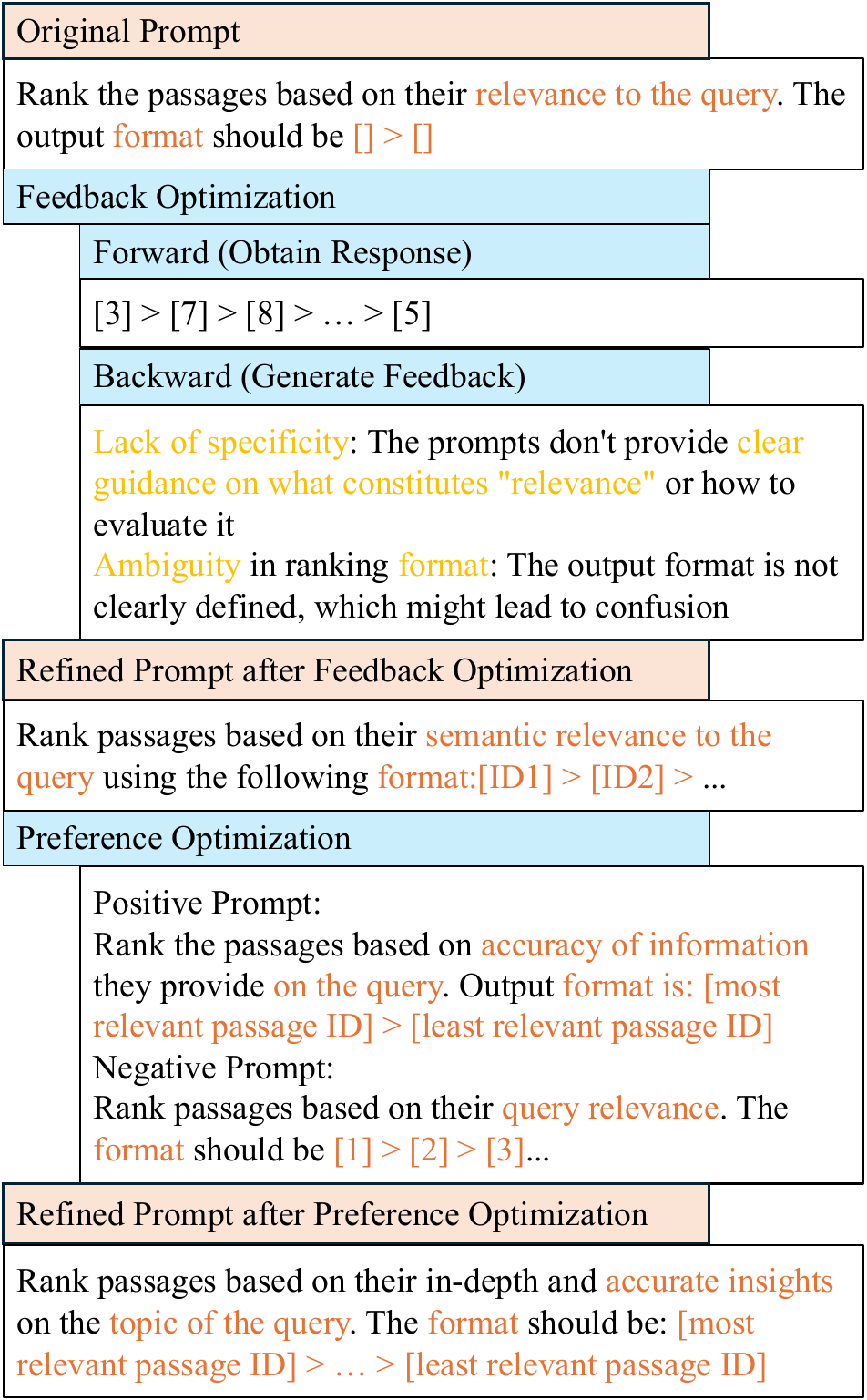}
    \caption{Illustration of \ours training responses. In Feedback Optimization, the LLM provides feedback on the original prompt and refine it based on the feedback. In Preference Optimization, the LLM mutate the refined prompt towards the positive prompt.}
    \label{figure_prompt_example}
\end{figure}

\paragraph{Qualitative Analysis.}
We provide qualitative examples of the training responses of \ours on LLaMA3. The illustration is shown in Figure \ref{figure_prompt_example}. During Feedback Optimization, the LLM provides feedback on the quality of the original prompt, such as noting "lack of specificity" and "ambiguity in format", and refines the prompt based on this feedback. In Preference Optimization, the LLM further refines the prompt based on preference alignment with the positive prompt while disfavoring the negative one. A new prompt that mutates the current prompt toward the positive prompt is then generated.

\section{Conclusion}
In this paper, we present a novel automatic prompt engineering algorithm named \ours for passage relevance ranking. \ours aims to reduce human effort in designing prompt for zero-shot LLM reranking and unlock the potential of prompt optimization. It iteratively generates refined prompts based on feedback optimization of current prompts and preference optimization using positive and negative prompt demonstrations. A comprehensive investigation using GPT4, GPT3.5, LLaMA3, and Qwen2, along with the widely acknowledged TREC and BEIR benchmarks, consistently demonstrates the performance improvements achieved by \ours. We further illustrate the transferability of prompts generated by \ours across diverse datasets and architectures. All investigations together indicate the effectiveness of the novel prompt preference optimization introduced in \ours.

\section{Acknowledgments}

This research is partially funded by research grants to Metaxas from NSF: 2310966, AFOSR 23RT0630, and NIH 2R01HL127661.
\bibliographystyle{ACM-Reference-Format}
\bibliography{sample-base}

\appendix
\section{Implementation Details}\label{appendix_implementation_details}
\paragraph{Detailed Information of Benckmarks} More detailed information about the number of queries and corpus for the test datasets is provided in Table~\ref{table_appendix_dataset_information}.

\begin{table}[hb]
\centering
\caption{Test Datasets Information}
\label{table_appendix_dataset_information}
\resizebox{0.5\textwidth}{!}{
\begin{tabular}{l|ccc}
\toprule
\textbf{Benchmark} & \textbf{Dataset} & \textbf{\#Queries} & \textbf{\#Corpus} \\ 
\midrule
\multirow{2}{*}{\textbf{TREC}} & DL19 & 43 & 8.8M \\
& DL20 & 54 & 8.8M \\
\midrule
\multirow{8}{*}{\textbf{BEIR}} & Covid & 50 & 171K \\
& NFCorpus & 323 & 3.6K \\
& Signal & 97 & 2.9M \\
& News & 57 & 595K \\
& Robust04 & 249 & 528K \\
& Touche & 49 & 382K \\
& DBPedia & 400 & 4.6M \\
& SciFact & 300 & 5K \\
\bottomrule
\end{tabular}}

\end{table}

\paragraph{Implementation Details of~\ours.} We construct our training set by sampling 100 queries from the MS MARCO v1 training split following Section \ref{section_build_training_dataset}. Our traning dataset is shared in the supplemental materials. We train \ours for three epochs, the batch size in Feedback Optimization is 1, and we utilzie the top 1 positive prompt and bottom 1 negative prompt in Preference Optimization.

\end{document}